\documentclass[conference]{IEEEtran}
\IEEEoverridecommandlockouts
\usepackage{cite}
\usepackage{amsmath,amssymb,amsfonts}
\usepackage{algorithmic}
\usepackage{graphicx}
\usepackage{textcomp}
\usepackage{xcolor}
\usepackage{booktabs}
\def\BibTeX{{\rm B\kern-.05em{\sc i\kern-.025em b}\kern-.08em
    T\kern-.1667em\lower.7ex\hbox{E}\kern-.125emX}}
\usepackage[hidelinks,colorlinks=true,linkcolor=blue,citecolor=blue]{hyperref}

\begin{document}

\title{Improving Graduate Outcomes by Identifying Skills Gaps and Recommending Courses Based on Career Interests}

\author{
\IEEEauthorblockN{Rahul Soni$^{1}$, Basem Suleiman$^{1}$, Sonit Singh$^{1}$}
\IEEEauthorblockA{\textit{$^{1}$School of Computer Science and Engineering, University of New South Wales, Sydney, Australia}\\
\textit{Email: \url{sonit.singh@unsw.edu.au}}
}}

\maketitle

\begin{abstract}
This paper aims to address the challenge of selecting relevant courses for students by proposing the design and development of a course recommendation system. The course recommendation system utilises a combination of data analytics techniques and machine learning algorithms to recommend courses that align with current industry trends and requirements. In order to provide customised suggestions, the study entails the design and implementation of an extensive algorithmic framework that combines machine learning methods, user preferences, and academic criteria. The system employs data mining and collaborative filtering techniques to examine past courses and individual career goals in order to provide course recommendations. Moreover, to improve the accessibility and usefulness of the recommendation system, special attention is given to the development of an easy-to-use front-end interface. The front-end design prioritises visual clarity, interaction, and simplicity through iterative prototyping and user input revisions, guaranteeing a smooth and captivating user experience. We refined and optimised the proposed system by incorporating user feedback, ensuring that it effectively meets the needs and preferences of its target users. The proposed course recommendation system could be a useful tool for students, instructors, and career advisers to use in promoting lifelong learning and professional progression as it fills the gap between university learning and industry expectations. We hope that the proposed course recommendation system will help university students in making data-drive and industry-informed course decisions, in turn, improving graduate outcomes for the university sector.
\end{abstract}

\begin{IEEEkeywords}
Course recommendation system, career interests, identifying skills gap, improving graduate outcomes, artificial intelligence, recommendation systems, large language models
\end{IEEEkeywords}

%%%%%%%%%%%%%%%%%%%%%%%%%%%%
%%%%%%%%%%%%%%%%%%%%%%%%%%%%
\section{Introduction}
Recommender systems have been increasingly popular in the last several years, and recommendation algorithms are now easily incorporated into a wide range of daily tasks. For example, platforms like Netflix routinely curate personalised movie suggestions under the `Top Picks for You' section, tailored to individual user interests. Often, these recommendations precisely match user preferences and enhance their viewing experience. Institutes of higher education provide students the skill set, preparing graduates to land into their dream career. However, the plethora of courses and learning objectives makes the selection of courses that are relevant to students' career interests a challenging task. The exhaustive list of elective courses with extensive details in the course outlines and limited time to make decision, makes the task of course selection a daunting task. Apart from this, with very limited chance of getting advice, either from peers or academics, can lead student to make suboptimal decisions~\cite{Fernandez:2017:The_paradox}. To overcome such challenges in higher education, a recommendation system, similar to online shopping or streaming platforms, can help students to make optimal decisions.  Inspired by this, our Course Recommendation System (CRS) endeavours to mirror such effectiveness by suggesting courses that align closely with industry standards. The motivation for this study is to close the gap between academia and industry, making sure that students acquire knowledge and skills as required by the industry. By developing a CRS, the goal is to make it easier for students to choose courses that would prepare them for their future employment.

%%%%%%%%%%%%%%%%%%%%%%%%%%%%
%%%%%%%%%%%%%%%%%%%%%%%%%%%%
\section{Related Work}
In this section, we provide a brief overview of recommender systems and related work on course recommendation system. 

\subsection{Recommender systems}
\iffalse
A number of factors ranging from students' career interests to major requirements and grade potential are considered for making the course recommendation system.  
\fi 

There are many recommendation system that incorporates collaborative filtering methods, content-based filtering methods, or the combination of them as a hybrid system.

\subsubsection{Content-based filtering} Content-based filtering focus on the properties of the items and recommends items based on their similarity to content a user has either explicitly or implicitly indicated a preference for in the past. The recommendations are computed based on the characteristics of each item and the user preferences for items. The item characteristics are stored in the item profile data structure while the user preferences are stored in the user profile data structure. The text is then evaluated based on the vector value and this process is called vectorisation. The content-based filtering then uses the vector space modal to measure the degree of similarity between the vectors of item and user preference. The smaller the angle between two vectors, the more similar they are. Term frequency-inverse document frequency (TF-IDF) serves as a prevalent method for extracting features from items. In this technique, each feature within the item and user profile vectors is depicted by its TF-IDF weight. These normalized vectors enable the comparison of item similarities with user preferences. Cosine similarity emerges as a widely utilized metric for assessing similarities between two vectors within the vector space model. In this method, for each item in the collection, the dot product of the item and user profile vectors is computed using Equation~\ref{eq:sim}.

\begin{equation} \label{eq:sim}
    \text{sim}(\vec{item}, \vec{user}) = \frac{\vec{item}}{\left | \vec{item} \right |} \cdot \frac{\vec{user}}{\left | \vec{user} \right |}
\end{equation}

\subsubsection{Collaborative filtering} Collaborative filtering is used to generate personalised recommendations for users based on their preferences, as well as the preferences of similar users. Unlike content-based algorithm, where the focus is on the items' attributes, collaborative filtering analyse user interactions with different items to find similarities and make recommendations based on this information. There are two main type of collaborative filtering, namely, \emph{user-based} and \emph{item-based}. The user-based algorithm recommends items to the target user by identifying the users who have similar preferences to the target user. For example, this algorithm will match a current student to another student who is sharing a similar pathway to them. It then recommends courses to the current student based on what the other student has already done. In contrast to the user-based algorithm, the item-based approach focuses on the similarities between the items rather than users. It recommends items similar to those that the user has previously liked. For example, item-based filtering uses students' grades in other subjects to recommend a course. Both, user-based and item-based, filtering algorithms uses the K-nearest neighbours (KNN) algorithm to compute the similarities between users or items. KNN is a distance based algorithm which is faster in the item-item similarity computation as compared to user-user similarity. This is because the number of items is usually smaller than the number of users in the system. Moreover, item-item similarity does not change overtime whereas the users' preferences can change frequently. Therefore, many systems uses item-based filtering for its efficiency and scalability.

\begin{figure}[htbp]
    \centering
    \includegraphics[width=1\linewidth]{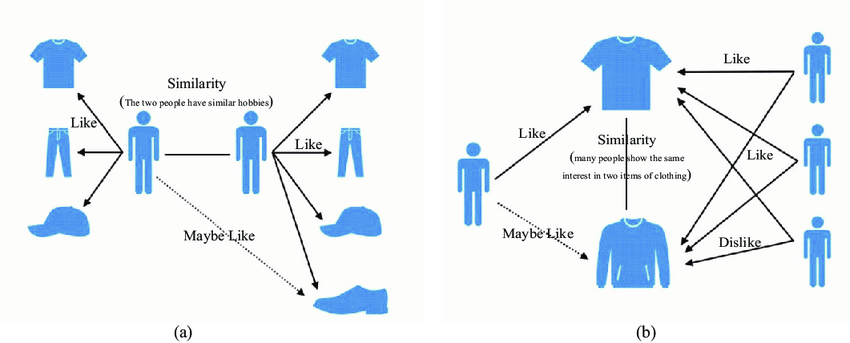}
    \caption{The collaborative filtering algorithms: (a) user-based; (b) item-based.~\cite{Collaborative_Filtering_Fig}}
\end{figure}

\subsubsection{Hybrid filtering} Hybrid filtering combines the two main filtering technique - Content-based and Collaborative. It combines the filtering technique to overcome the limitations of each individual methods and provide more accurate and diverse recommendations. It approaches the problem in various ways such as making content-based and collaborative recommendations separately and then combining them or by adding one algorithm to the other. 

\subsection{Course recommendation systems}
Few studies have looked at the development of recommender systems that take into account many influencing elements in recent years~\cite{Frej:2024:course_recommendation, Cha:2024:the_impact}. Shabab \emph{et al.,}~\cite{Shahab:2019:Next_Level_Course_Recommender_System} presented a course recommender system that intends to further enhance the quality of career preparation via assisting in placing educational courses in marketable skills required within the workforce. The system uses the latest content-based filtering algorithms combined with k-means clustering and TF-IDF keyword extraction to analyse job descriptions. It then goes further to recommend to you related courses. This review gave much emphasis to the promising results towards matching educational offerings to the career interest of the students, indicating its potential in assisting students towards meeting the current demands of the job market. This system effectively acts as a bridge between academic learning and the need of the market with its updated course recommendations in the light of current market trends, which will automatically enhance employability for the students post their graduation. However, its precise clustering mechanism apparently needs further fine-tuning, maybe to provide it with an improved correct way of providing recommendations regarding an appropriate course, pointing to the possible limitations of the effectiveness of its current method in varied scenarios. Wang \emph{et al.,}~\cite{Wang:2020:employee_training} proposed to improve employee training by matching courses to individual career goals with a personalised course recommender system named Demand-aware Collaborative Bayesian Variational Network (DCBVN). The system adopts a unique combination of data on employee competency and career aspiration data through advanced modelling techniques, resulting in personalised and explainable course recommendations. Therefore, real-world data on DCBVN was tested. It works well in dealing with the problem of data sparsity and the challenge for new-user recommendations and thus recommends the utility of strategic talent management. The system DCBVN suggests, therefore, is highly individualised and explainable by the inclusion of the current competencies of the employee and career aspirations and offers a solution to individual and organisational training needs more effectively. However, the performance of the system depends heavily on the precision and comprehensiveness of the profile of skills provided by the employees. This could have posed a limitation in environments where data collection was inconsistent or incomplete.

%%%%%%%%%%%%%%%%%%%%%%%%%%%%
%%%%%%%%%%%%%%%%%%%%%%%%%%%%
\section{Methodology}
This study aims to design and implement a course recommendation system that can recommend courses based on user's career interests and as per the recent trends in industry.

\subsection{Dataset sources}
The initial step is to gather valuable data from various sources that are related to the course recommender system. There are two sources of data that we use - the course database and the job description database.

\subsubsection{Course data}
This dataset contains information related to the courses such as name, description, and metadata. The dataset include both undergraduate and postgraduate courses with each course containing the Course ID, Course Name and Course Description. The important course description can be found in the Course Learning Outcome (CLO) section for each course. 

\subsubsection{Jobs data}
This dataset contains comprehensive details related to the job postings. This information can be easily found on various job listings websites such as - LinkedIn, Seek, and Indeed. These platforms offer the description that can easily be crawled.

\subsection{System Architecture}
The overall system architecture for our course recommender system will consist of the following structure:
\begin{enumerate}
    \item \textbf{Frontend Implementation:} The front-end system that will accept users' requests.
    \item \textbf{Backend Implementation:} The data will be processed using one of the text processing algorithm and stored in a data-lake. 
    \item \textbf{Recommendation Engine:} Compute the recommendation using the hybrid filtering algorithm. 
\end{enumerate}

After research, the following libraries and frameworks are used in the development of the course recommendation system.

\begin{table}[]
    \centering
    \caption{Table showing the technical stack used}
    \begin{tabular}{l|l}
    \toprule
    \textbf{Component} & \textbf{Framework} \\
    \midrule
    UI Element & Figma, React, Javascript \\
    Keyword Extraction & Scikit-learn, RAKE-NLTK \\
    Text Processing & NLTK, Pandas, Bs4 \\
    \bottomrule
    \end{tabular}
\end{table}

\subsubsection{Data Preprocessing Engine} After finding all the relevant data sources, it is important to perform data cleaning and processing so that the gathered data can be used effectively in our course recommendation system. Some basic data normalisation technique include: 
\begin{itemize}
    \item removing HTML tags
    \item removing punctuation
    \item removing excessive white space
    \item removing symbols
    \item using standardised vocabulary
    \item expand contractions
    \item lowercase and lemmatise the text
\end{itemize}

To enhance the system's performance, data processing was conducted independently, ensuring that only refined and optimised data was fed into the CRS system.

\subsubsection{HTML Tags} When the data is scraped from web sources such as LinkedIn, Indeed, and Seek, it is common to find the text data to be embedded with HTML tags. However, these tags are irrelevant for our recommender system. Therefore, it is important to remove all HTML tags.

\subsubsection{Punctuation} Punctuation symbols such as periods, commas, and question marks are only used for grammatical purposes but are not really helpful for the text analysis. Therefore, these symbols are usually removed from the text data.

\subsubsection{Excessive White Space} Symbols that can be safely eliminated include monetary symbols, mathematical symbols, and special characters as they may not provide significant value to text analysis tasks. In this preparation stage, the text's symbols are removed.

\subsubsection{Symbols} Sometimes the text data can contain unnecessary white space such as extra spaces, tabs or newline characters. They can hinder with the text analysis and it is better to just remove them during the text preprocessing.

\subsubsection{Contractions} Words that are reduced by joining two words with an apostrophe are called contractions. Examples of these include "don't" for "do not" and "can't" for "cannot." To maintain consistency across the text, expanding contractions entails substituting the contracted forms with their full-word equivalents.

\subsubsection{Lowercase and Lemmatising} To maintain consistency in case sensitivity, all text is converted to lowercase through the process of lowercasing. Lemmatization is the language process of breaking down words into their most basic form, or lemma. This makes it easy to analyse text by standardising terms and reducing inflectional forms to their common base form.

\subsubsection{Keyword Extraction Methods} We make use of different keyword extraction methods that helps us to gather data efficiently and also process it.

\begin{enumerate}
    \item \textbf{TF-IDF}: It is a statistical measure widely used in natural language processing and information retrieval to evaluate the importance of a word in a document relative to a collection of documents, known as a corpus. It aims to highlight words that are both frequent within a specific document (term frequency (TF)) and rare across all documents in the corpus (inverse document frequency (IDF)), thereby emphasising words that are unique and characteristic of the document's content.

    \begin{equation}
        \text{TF}(t, d) = \frac{\text{Number of occurrences of term } t \text{ in document } d}{\text{Total number of terms in document } d}
    \end{equation}
    
    \begin{equation}
        \text{IDF}(t, D) = \log{\frac{N}{\text{Number of documents containing term } t}}
    \end{equation}

    \begin{equation}
        \text{TF-IDF}(t, d, D) = \text{TF}(t, d) \times \text{IDF}(t, D)
    \end{equation}
    
    \item \textbf{RAKE}: The Rapid Automatic Keyword Extraction Algorithm (RAKE)~\cite{NER_RAKE} is a simple and efficient algorithm that extracts keywords or key phrases from text. It is especially effective in text mining and natural language processing applications where finding key phrases is critical. RAKE is noted for its simplicity, efficiency, and language independence, making it ideal for a variety of text analysis jobs. It does not rely on complicated language models or training data, making it ideal for situations where speed and ease of deployment are critical. Each keyword's score is calculated by adding the scores of its constituent terms. They employ measures like word degree, frequency, and ratios. RAKE is known for its speed and precision when compared to TextRank. It is also excellent in creating keyword phrases.

    \item \textbf{TextRank}: It is a graph-based ranking algorithm primarily used for extractive summarization by identifying and extracting key sentences from a document. It extends the principles of the PageRank algorithm, originally developed to rank web pages based on their significance in search engines like Google. In TextRank, the text is represented as a graph where sentences serve as nodes, and relationships between sentences (e.g., similarity or semantic links) are represented as edges. Each node is scored based on its connections to other nodes in the graph. The algorithm prioritizes sentences that are highly connected to other significant sentences, effectively identifying the most relevant sentences for summarization. While TextRank is also capable of extracting keywords, its primary focus is on summarizing documents by highlighting the most critical sentences.

    \item \textbf{SpaCy}: It is a modern Python library for advanced NLP, it gives efficiency and ease of use in handling large text datasets.
    
    \item \textbf{Word Cloud}: It is a visual representation of text data where the size of each word represent it frequency or importance in the document.
    
    \item \textbf{KeyBERT}: It is a keyword extraction technique that uses BERT embeddings for the identification of the most relevant keywords and phrases within a document.
    
    \item \textbf{YAKE}: It is a lightweight algorithm for automatic keyword extraction that relies on statistical data from the texts to identify the most significant keywords.
\end{enumerate}

\subsubsection{Learning Engine} The Learning Engine is crucial in identifying patterns and structures in the dataset. It can use advanced algorithms to make this possible. One approach used is the K-Means Clustering algorithm. It is very efficient and scalable which makes it ideal for dealing with large datasets. The algorithm divides the data into separate groups, also known as clusters, using the similarity metrics between the data. The algorithms key aspects include:
\begin{itemize}
    \item There are always \(k\) clusters
    \item Each cluster has at least one item
    \item There is no cluster overlapping
\end{itemize}

\subsubsection{Recommender Engine} The recommender engine uses an hybrid of the content-based and the collaborative filtering to recommend relevant courses to the students. Unlike traditional methods of just focusing on the item-item or user-user interaction, this hybrid model will utilise a combination of both user-item interaction for enhanced recommendations. Similarities between items can be determined by the use of the vector space model. 

\subsection{System Evaluation}
It is very important to assess the performance of the Course Recommendation System (CRS) in such a way that its effectiveness and usefulness for giving recommendations of courses commensurate with the career goal and academic needs of the student are guaranteed. This section presents proposed evaluation methodologies to be employed in the rigorous testing and human assessment of the system on completion.

\iffalse
\subsubsection{Accuracy Metrics}
\begin{itemize}
    \item \textbf{Precision and Recall:} It is an accuracy metric for recommendations made on the system. Precision finds the proportion of recommended courses considered relevant, while recall looks at the proportion of recommended relevant to all the relevant courses.
    \item \textbf{F1-Score:} It is one of the F1-measure metrics balancing both precision and recall of the CRS to have a single score for a representative score for the two accuracy.
\end{itemize}
\fi 

\subsubsection{Usability Testing}
\begin{itemize}
    \item \textbf{User Satisfaction Surveys:} These would be post-interaction surveys with customers, aiming to measure customers' perceptions of several elements, such as ease of use of the CRS, design of the interface, and overall satisfaction with the course recommendations.
    \item \textbf{System Usability Scale:} This is one of the standard questionnaires being accepted across the globe for checking usability in order to provide reliable measuring tools, as per the users.
\end{itemize}

\subsubsection{Performance Evaluation}
\begin{itemize}
    \item \textbf{Response Time:} Measurement of Time through the system from Recommendations back and making a continuous assessment of the efficiency.
    \item \textbf{Scalability Tests:} are those that determine the size of data and how much the capacity of the system can grow without degradation of performance under the given loads and in the current environment of use.
\end{itemize}

\subsubsection{Criteria for Success}
\begin{itemize}
    \item The system should exhibit recommendations with high precision and recall.
    \item User feedback is expected to register high satisfaction with the recommendations made by the system and its usability level.
    \item The CRS should be efficient, working under variable loads.
\end{itemize}

\subsubsection{Tools and Resources}
\begin{itemize}
    \item Evaluate system metrics using tools for data analysis.
    \item User Feedback through Survey Software.
    \item Performance monitoring tools for system responsiveness and efficiency. Through these, the CRS project ensures that the system realises technical requirements when implemented, but with a high level of customers' satisfaction, practical utility, and end-users' satisfaction in their respective learning environments.
\end{itemize}

%%%%%%%%%%%%%%%%%%%%%%%%%%%%
%%%%%%%%%%%%%%%%%%%%%%%%%%%%
\section{Design \& Implementation}
In this section, we provide details about tools and frameworks used, design process evolving from low-fidelity wireframes to high-fidelity design, and finally, the backend development. The technology stack for the CRS was carefully selected to ensure scalability, flexibility, and efficiency across both the frontend and back-end. Each component played a crucial role in building a system that could process data effectively and deliver accurate recommendations in real time. 

\subsection{Frontend}
The frontend was designed to provide an intuitive and user-friendly interface for students. The tools and frameworks used for the frontend include:

\subsubsection{Figma} We use Figma for creating both low-fidelity and high-fidelity prototypes, ensuring iterative design improvements based on user feedback. It allowed user-testing to refine the UI/UX design.
  
\subsubsection{React} We choose React framework because of its component-based architecture and scalability. The react framework ensured reusable UI components, making the development process efficient and modular.

\subsubsection{JavaScript} It is the to-go choice as programming language for front-end developers and enhancing the user interface with seamless interactivity and dynamic behaviour.

\subsection{Backend}
The backend supported data processing, storage, and recommendation generation. It consisted of the following technologies:

\subsubsection{Firebase} We use firebase real-time database management, user authentication, and cloud storage of JSON files. It also simplifies integration with other Google services, such as the RAG model.

\subsubsection{Google Cloud RAG Model} The Retrieval Augmented Generation (RAG) model played a key role in recommendation generation by identifying gaps in user skills and matching them with relevant courses. The RAG capabilities ensured personalized, context-aware course suggestions.

\subsection{APIs}
The technology stack for the Course Recommendation System (CRS) included specialised tools and APIs for efficient data processing and skill extraction.

\subsubsection{Lightcast API } The Lightcast API automatically recognise relevant skills and competencies in the textual data. It seamlessly integrates with the back-end, processing large volumes of text data with minimal latency. 

\subsubsection{Open Skills API } It uses the power of the Open Skills API to help find useful and in-demand skills in the resumes.

\subsubsection{PyPDF2 } We use PyPDF2 parser for extracting raw text from resumes in PDF format. The preprocessing pipeline provided input for downstream processes like skill extraction via Lightcast.

\subsection{Development Tools}
To streamline the development workflow, additional tools were employed:

\subsubsection{GitHub} Managed version control and keep my data secure.
\subsubsection{Visual Studio Code } Provided an integrated development environment (IDE) for writing and debugging code.
\subsubsection{Postman} Assisted in testing API endpoints for data retrieval and recommendation validation.

\begin{figure*}[]
    \centering
    \includegraphics[scale=0.29]{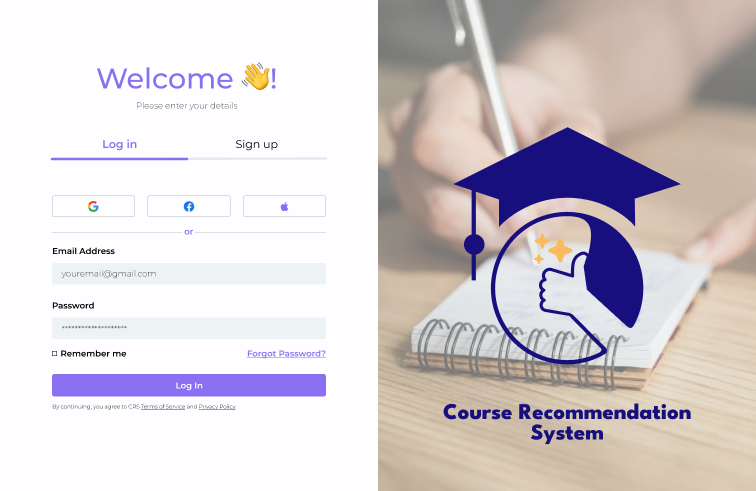}
    \caption{Login/Sign up page for the course recommendation system.}
    \label{fig:sign-up}
\end{figure*}

\begin{figure*}[]
    \centering
    \includegraphics[scale=0.29]{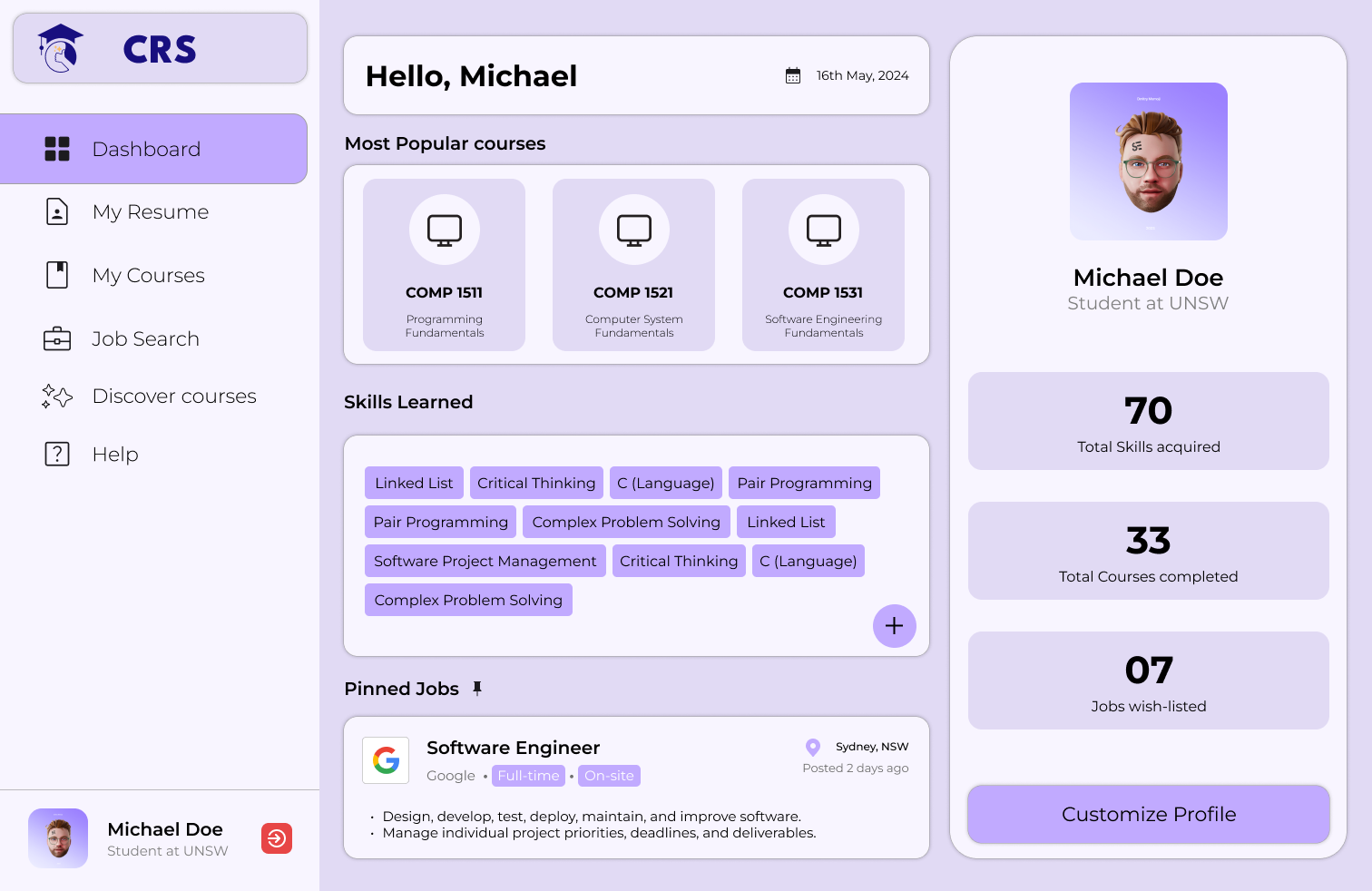}
    \caption{Dashboard/Home page for the course recommendation system.}
    \label{fig:dashboard}
\end{figure*}

\begin{figure*}[]
    \centering
    \includegraphics[scale=0.29]{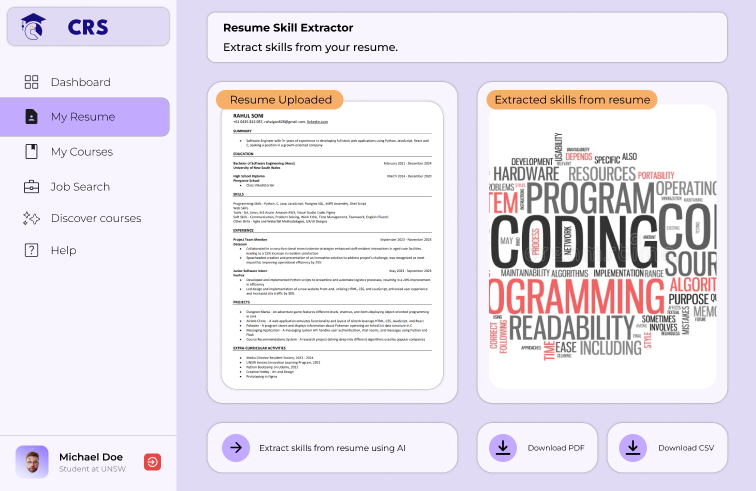}
    \caption{Upload Resume page for the course recommendation system.}
    \label{fig:resume}
\end{figure*}

\begin{figure*}[]
    \centering
    \includegraphics[scale=0.29]{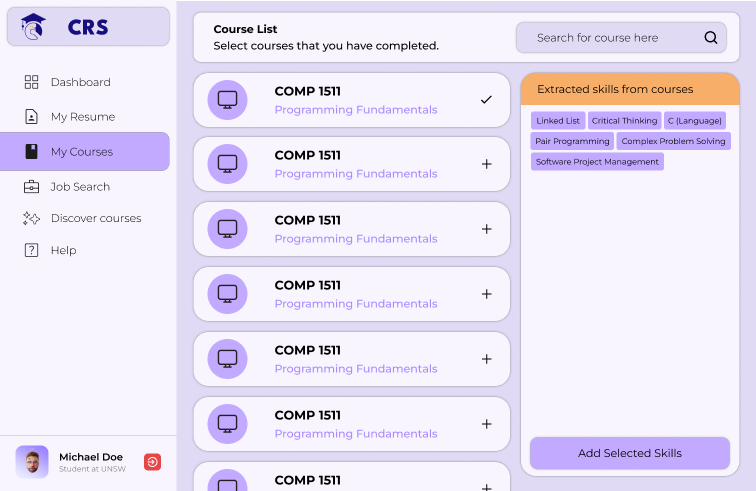}
    \caption{Course selection page for the course recommendation system.}
    \label{fig:course-list}
\end{figure*}

\begin{figure*}[]
    \centering
    \includegraphics[scale=0.29]{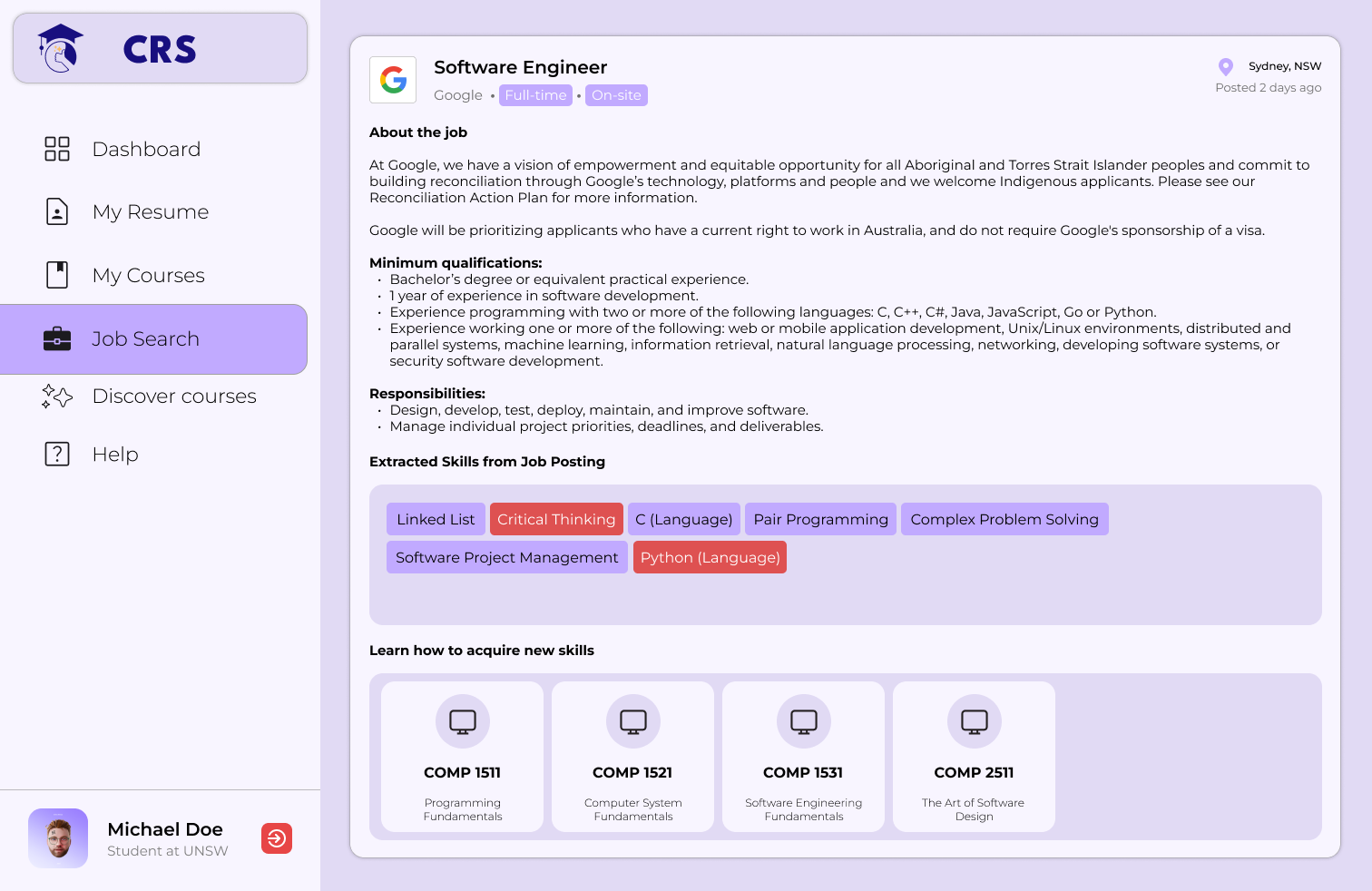}
    \caption{Job postings page for the course recommendation system.}
    \label{fig:job-postings}
\end{figure*}

\begin{figure*}[]
    \centering
    \includegraphics[scale=0.29]{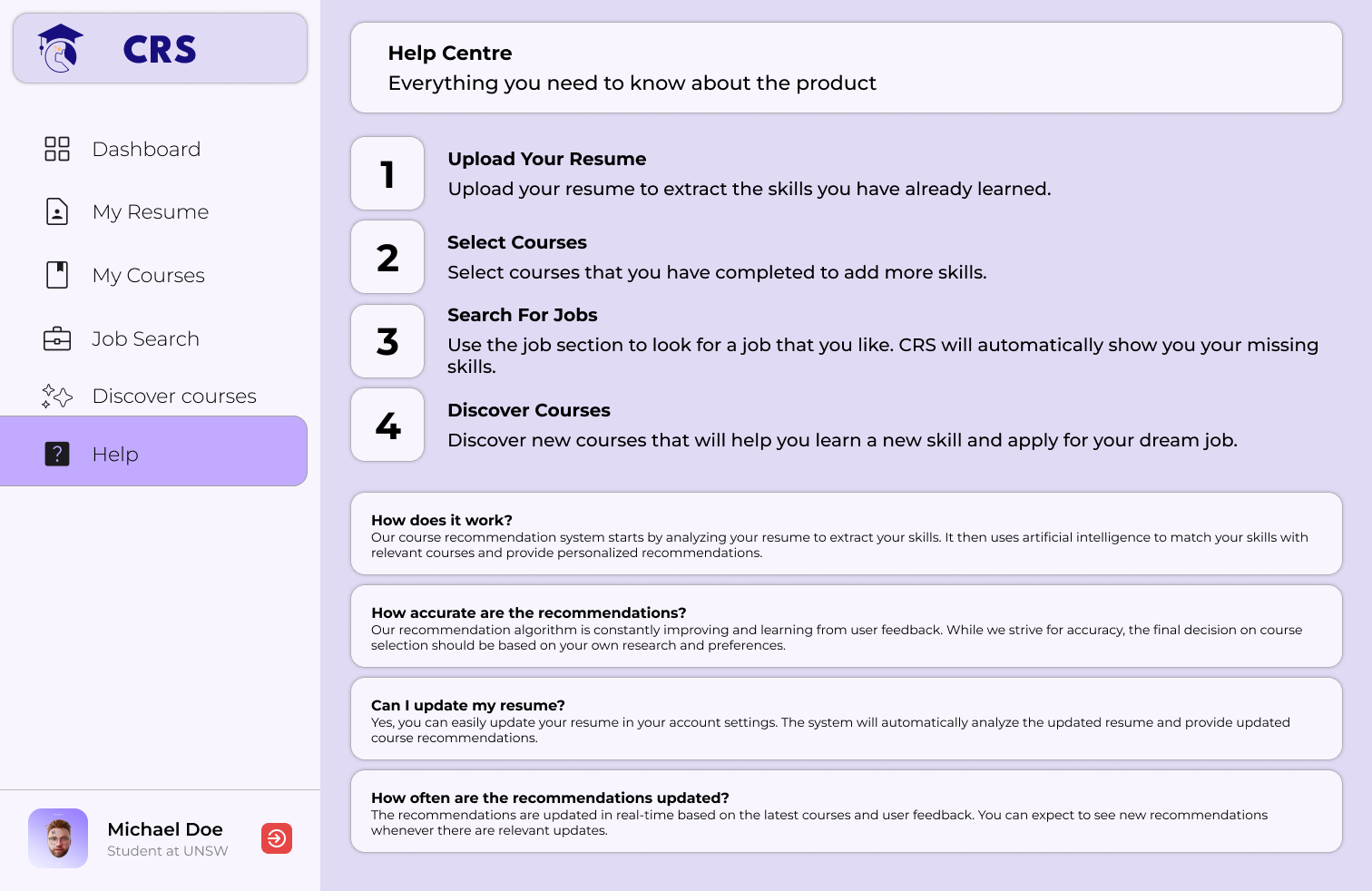}
    \caption{Help Section page for the course recommendation system.}
    \label{fig:help-section}
\end{figure*}

\iffalse
\subsection{Key Design Considerations}
Building on the foundation of low-fidelity designs, high-fidelity prototypes were developed to provide a realistic view of the user interface. These prototypes:

\begin{itemize}
    \item \textbf{Simplicity:} Prioritized a clean, intuitive interface to ensure users could easily understand and navigate the system.
    \item \textbf{Consistency:} Maintained a uniform design language throughout the application, enhancing the overall experience.
\end{itemize}
\fi 

%%%%%%%%%%%%%%%%%%%%%%%%%%%%
%%%%%%%%%%%%%%%%%%%%%%%%%%%%
\section{System Architecture and Components}
\begin{figure*}[]
    \centering
    \includegraphics[scale=0.1]{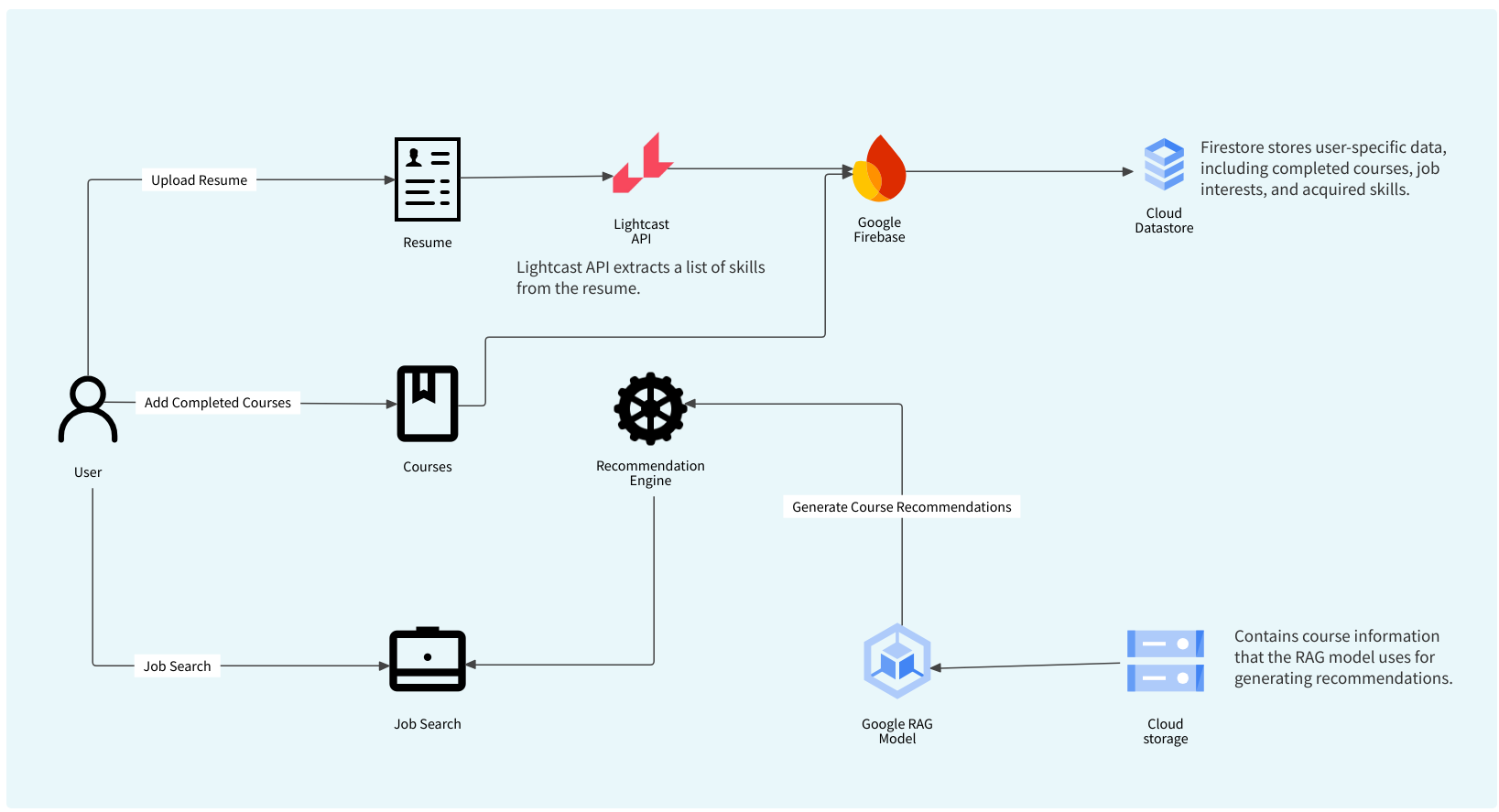}
    \caption{System Flow Diagram of the Course Recommendation System.}
    \label{fig:system_diagram}
\end{figure*}

The system architecture of the CRS was designed to efficiently handle data processing, skill analysis, and recommendation generation. Figure~\ref{fig:system_diagram} provides system flow diagram for the CRS. It consists of three primary components: the frontend, backend, and recommendation engine. Each component plays a vital role in ensuring a seamless and reliable user experience.

\subsection{Frontend}
The frontend served as the user-facing layer of the system, providing an intuitive interface for interaction. We implemented frontend using React and styled with custom CSS, ensuring responsiveness across multiple devices. Key features for \emph{user input} include:

\begin{itemize}
    \item options for uploading resumes in PDF format or directly selecting courses for skill analysis.
    \item a clean and organised dashboard for navigating course recommendations and skill gaps.
\end{itemize}

\subsection{Backend}
The backend handled data processing and interaction with external services. It included the following components:

\begin{itemize}
    \item The Lightcast API extracted skills from job descriptions, resumes, and course metadata, providing standardized skill sets for recommendation matching.
    \item The PyPDF2 processed uploaded PDF resumes to extract text for downstream skill analysis.
\end{itemize}

\subsection{Google Cloud}
The recommendation engine is the core of the system which is integrated with Firebase and Google Cloud to dynamically process user data and update recommendations instantly.

\begin{itemize}
    \item Firebase Integration managed user authentication, real-time database storage, and cloud storage for user-uploaded files and processed data.
    \item Google Cloud RAG model used for generating personalized recommendations based on missing skills identified through Lightcast's output.
    \item Offered retrieval-augmented responses tailored to the user’s career goals and current skill set.
\end{itemize}

\subsection{Data Flow}
The architecture follows a streamlined data flow:
\begin{enumerate}
    \item \textbf{User Input:} Upload Resumes or select completed courses.
    \item \textbf{Data Preprocessing:} Text is extracted, cleaned, and formatted using PyPDF2 and preprocessing libraries.
    \item \textbf{Skill Extraction:} Lightcast API identifies key skills and competencies.
    \item \textbf{Recommendation Generation:} The Google Cloud RAG model processes missing skills and provides targeted course suggestions.
    \item \textbf{Output:} Recommendations are displayed on the frontend with actionable insights.
\end{enumerate}

\subsection{Data Processing}
Data preprocessing is a critical step in the CRS as it ensures that raw data is transformed into a clean, structured, and analysable format. The system processes various data types, including resumes, job descriptions, and course metadata, to extract meaningful skills and competencies.

\subsubsection{Data sources} The CRS relies on three main data sources:
\begin{enumerate}
    \item \textbf{Resumes:} Uploaded by users to identify their existing skills.
    \item \textbf{Job Descriptions:} Scraped from job boards like LinkedIn, Indeed, and Seek and stored in database. Optional uploaded by users to determine industry skill demands.
    \item \textbf{Course Metadata:} Includes course descriptions, learning outcomes, and associated skills provided by UNSW.
\end{enumerate}

\subsubsection{Skill Extraction} To identify relevant skills, the CRS uses the Lightcast API, which extracts a list of recognized skills from the cleaned text. It matches extracted skills to industry-defined standards, ensuring alignment with market demands.

\subsection{Firebase Integration}
The integration of Firebase and Google Cloud RAG was critical to enabling real-time data processing, secure storage, and personalized course recommendations in the CRS. 

\subsubsection{Key Features}: Firebase provided a robust backend infrastructure for managing user data, ensuring scalability and security throughout the system. Key features utilised include \emph{authentication} as Firebase handled secure login and account management for users. Also, it supported multiple authentication methods (e.g., email/password, Google Sign-In). The Cloud Firestore stored user profiles, skills extracted, and course data. It also enabled real-time synchronisation of data, ensuring that updates were instantly reflected across devices. Finally, the cloud storage ensured easy retrieval and integration with other backend components, such as the RAG model, and also stored uploaded resumes and course data from UNSW. The entire setup can automatically be scaled to handle large number of users and data requests without any performance degradation.

\subsection{Learning Engine (RAG)}
The learning engine and the recommendation logic form the core of the Course Recommendation System (CRS). This component processes extracted skills, identifies skill gaps, and generates personalized course recommendations using the Google Cloud's RAG Model. The Google Cloud RAG (Retrieval-Augmented Generation) model was used to enhance the recommendation process.

\subsubsection{Example Workflow}
\begin{enumerate}
    \item \textbf{Input Data:} A user uploads a resume and select completed courses at UNSW.
    \item \textbf{Skill Extraction:} Lightcast API identifies existing skills (e.g., "Python," "SQL").
    \item \textbf{Comparison:} Identifies missing data by comparing input data with the results of the job skills.
    \item \textbf{Output:} This highlights skill gaps which form the basis for recommendation generation.
\end{enumerate}

%%%%%%%%%%%%%%%%%%%%%%%%%%%%
%%%%%%%%%%%%%%%%%%%%%%%%%%%%
\section{Evaluation and Testing}
\begin{figure}[]
    \centering
    \includegraphics[width=1\linewidth]{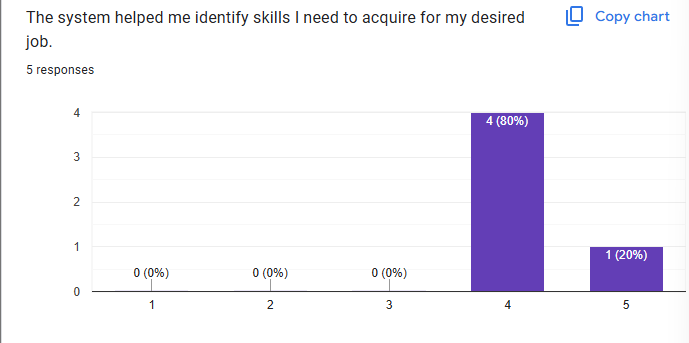}
    \caption{Results of the survey providing how skill extraction was helpful.}
    \label{fig:system_eval_1}
\end{figure}

\begin{figure}[]
    \centering
    \includegraphics[width=1\linewidth]{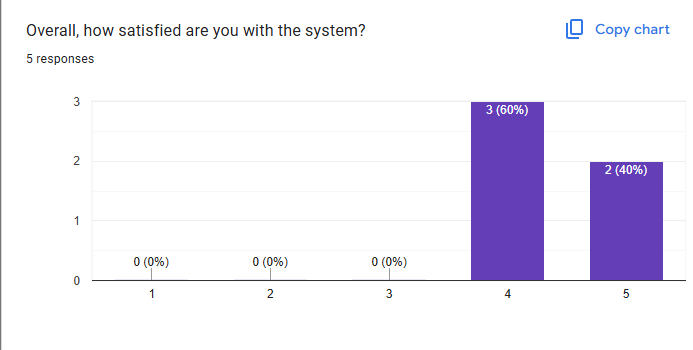}
    \caption{Results of the survey providing user's overall system satisfaction.}
    \label{fig:system_eval_2}
\end{figure}

Evaluation and testing are critical phases in ensuring the reliability, efficiency and user-friendliness of the CRS. This section details the methods and criteria used to validate the system's functionality, performance, and overall user experience. Figure~\ref{fig:system_eval_1} and Figure~\ref{fig:system_eval_2} shows results of responses for CRS's functionality of skills extraction and overall satisfaction rate of the user. 

\subsection{Usability Testing}
Usability testing focused on evaluating the system's design and interaction flow:
\begin{enumerate}
    \item \textbf{User Satisfaction Surveys}: We collected feedback from users on ease of use, navigation, and overall satisfaction with the CRS. The questions were on Likert scale (e.g., "How satisfied are you with the recommendations provided?").
    
    \item \textbf{System Usability Scale}: We also apply a globally accepted usability testing tool that provided a standardised score based on user responses.

    \item \textbf{Task Completion Rates}: Finally, we measured the percentage of users able to complete key tasks, such as uploading resumes, analysing skill gaps, and accessing recommendations.
\end{enumerate}

%%%%%%%%%%%%%%%%%%%%%%%%%%%%
%%%%%%%%%%%%%%%%%%%%%%%%%%%%
\section{Limitations and future work}
The development of the course recommendation system involved addressing various technical, design, and operational challenges. Our current CRS focuses on extracting technical skills (\emph{e.g.,} Python, SQL, JavaScript) and does not capture soft skills (\emph{e.g.,} teamwork, communication, problem-solving) that are critical for holistic career development. The future work aims to expand the skills extraction capability tailored for soft skill analysis. Also, current system lacks any accessibility features. The system needs to accommodate users across diverse devices and varying accessibility needs. These problem can be easily overcome by incorporating accessibility features, such as keyboard navigation and high-contrast mode.

%%%%%%%%%%%%%%%%%%%%%%%%%%%%
%%%%%%%%%%%%%%%%%%%%%%%%%%%%
\section{Conclusion}
In this paper, we designed and developed a course recommendation system, developed
to bridge the gap between academic offerings and industry needs. The system success-
fully provides personalized course suggestions that align with students’ career goals and
current job market trends. By using machine learning, data analytics, and cutting-edge
APIs, the course recommendation system helps users identify skill gaps and suggests relevant courses to fill them. The system incorporates the Lightcast API for skill extraction and the Google Cloud RAG model for generating course recommendations, ensuring both precision and flexibility. It has been thoroughly tested for usability, performance, and accuracy, with impressive results in user satisfaction and scalability. This system’s successful implementation showcases its potential as a powerful tool for students, educators, and career advisors navigating the constantly evolving job market. This work not only proves the concept’s viability but also sets the stage for future advancements in personalized learning and career development tools.

%%%%%%%%%%%%%%%%%%%%%%%%%%%%
%%%%%%%%%%%%%%%%%%%%%%%%%%%%
\section*{Acknowledgment}

This research was supported by Katana, the high performance computing facility at the University of New South Wales. The authors also  acknowledge the financial support provided by the School of Computer Science and Engineering for API and cloud services used in the development of the course recommendation system.

\bibliographystyle{IEEEbib}
\bibliography{references.bib}

\end{document}